\documentclass[journal]{IEEEtran}
\usepackage{amsmath,amssymb,amsfonts}
\usepackage{algorithmic}
\usepackage{graphicx}
\usepackage{textcomp}
\usepackage{xcolor}
\usepackage{bm}

\begin{document}
	
	\title{Modern Neuromorphic AI: From Intra-Token to Inter-Token Processing}

    \author{Osvaldo Simeone, \IEEEmembership{Fellow,~IEEE}
\vspace{-0.5cm} 
\thanks{The author is with the Intelligent Networked Systems Institute (INSI),  Northeastern University London, One Portsoken Street, London E1 8PH, United Kingdom   (email: o.simeone@northeastern.edu). This work was supported  by the European Research Council (ERC) under the European Union’s Horizon Europe Programme (grant agreement No. 101198347), by~an Open Fellowship of the EPSRC (EP/W024101/1), and~by the EPSRC project (EP/X011852/1).}
}

	\maketitle
	
	\begin{abstract}
The rapid growth of artificial intelligence (AI) has brought novel data processing and generative capabilities but also escalating energy requirements. This challenge motivates renewed interest in neuromorphic computing principles, which promise brain-like efficiency through discrete and sparse activations, recurrent dynamics, and non-linear feedback. In fact, modern AI architectures increasingly embody neuromorphic principles through heavily quantized activations, state-space dynamics, and sparse attention mechanisms. This paper elaborates on the connections between neuromorphic models, state-space models, and transformer architectures through the lens of the distinction between intra-token processing and inter-token processing. Most early work on neuromorphic AI was based on spiking neural networks (SNNs) for intra-token processing, i.e., for transformations involving multiple channels, or features, of the same vector input, such as the pixels of an image. In contrast, more recent research has explored how neuromorphic principles can be leveraged to design efficient inter-token processing methods, which selectively combine different information elements depending on their contextual relevance. Implementing associative memorization mechanisms, these approaches leverage state-space dynamics or sparse self-attention. Along with a systematic presentation of modern neuromorphic AI models through the lens of intra-token and inter-token processing, training methodologies for neuromorphic AI models are also reviewed. These range from surrogate gradients leveraging parallel convolutional processing to local learning rules based on reinforcement learning mechanisms.
	\end{abstract}
	
	\section{Introduction}

\begin{figure*}[!t]
	\centering
	\includegraphics[width=0.8\linewidth]{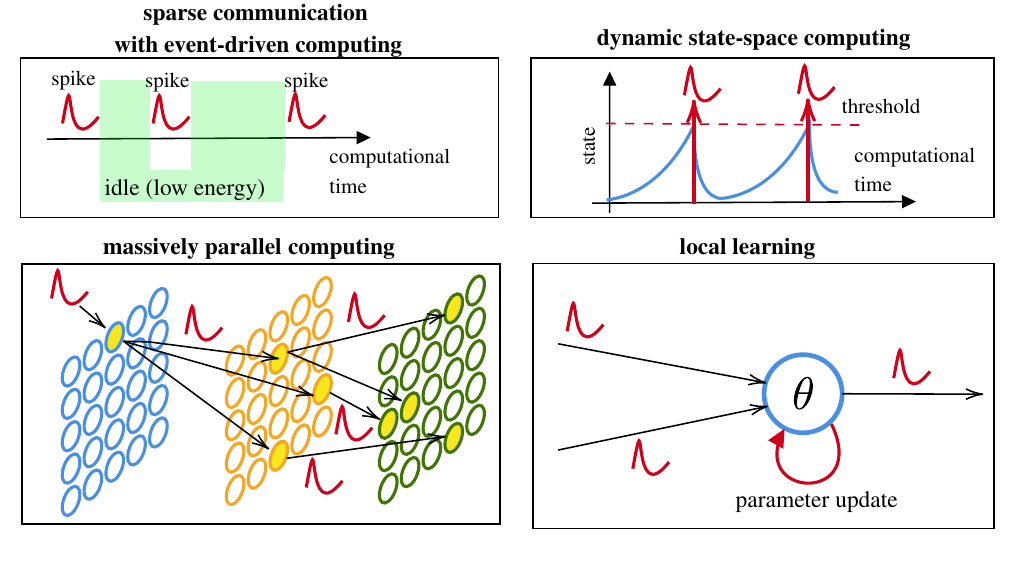}
	\caption{Over more than three decades of research, {neuromorphic engineering} has identified a number of principles from neuroscience that may be responsible for the energy efficiency of the brain, including:  \textbf{sparse} communication through discrete \textbf{spikes} (i.e., binary signals) with \textbf{event-driven} processing, \textbf{dynamic state-space} computations in neuronal membranes, analog  in-memory computing, massively \textbf{parallel} computing, and \textbf{local} learning rules.}
	\label{fig:neuroAI}
\end{figure*}

Modern artificial intelligence (AI) tools are widely believed to have the potential to transform our economies and societies, yet this progress comes at a well-documented steep cost in terms of energy resources for both training \cite{cottier2024rising} and inference \cite{smith2025aienergy}. The quest for a large \textbf{``intelligence-to-joule ratio''} \cite{sambanova2025intelligence} has long been the objective of \textbf{neuromorphic engineering}, which takes inspiration from the most energy-efficient intelligent system known: the human brain \cite{mead2020we}. Over more than three decades of research, {neuromorphic engineering} has identified a number of principles from neuroscience that may be responsible for the energy efficiency of the brain \cite{davies2021advancing,rajendran2025}, including (see Figure \ref{fig:neuroAI}):  \textbf{sparse} communication through discrete \textbf{spikes} (i.e., binary signals) with \textbf{event-driven} processing, \textbf{dynamic state-space} computations in neuronal membranes, analog  in-memory computing, massively \textbf{parallel} computing, and \textbf{local} learning rules.

\begin{figure}[!t]
\centering
\includegraphics[width=1\linewidth]{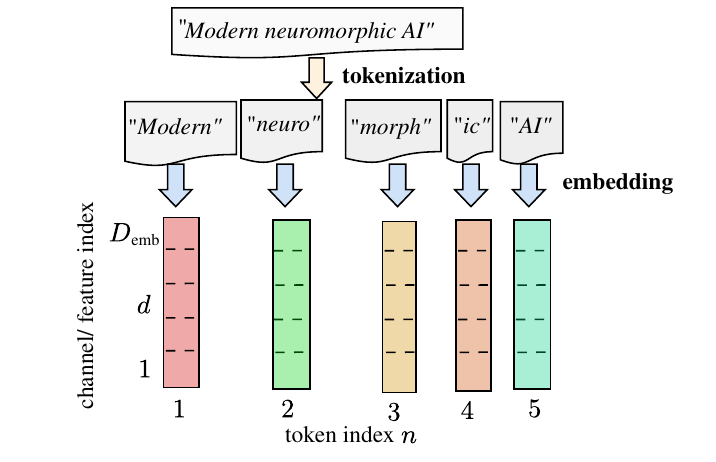}
\caption{Modern AI systems pre-process text -- and similarly other types of data --  via tokenization and embedding. These steps produce a matrix $\bm E$ with  $N$ columns (here $N=5$), each corresponding to a token embedding  containing $D$ channels.}
\label{fig:tokenization}
\end{figure}

Recent developments in mainstream AI have incorporated a number of lessons from neuromorphic engineering \cite{rajendran2025}. Most notably, state-of-the-art large language models (LLMs) have been demonstrated with activations quantized to integers with resolutions as low as 4 bits \cite{ma2024era,liu2023llm}  -- moving towards the neuromorphic regime of spike-based (i.e., binary)  communication. Furthermore,  some of the best-performing sequence-to-sequence models are designed based on state dynamics \cite{gu2021efficiently,gu2023mamba} -- thus reproducing the stateful operation of typical neuromorphic models. This convergence reflects fundamental computational principles: efficient systems must exploit temporal structure if present and avoid unnecessary computation through sparsity.

Despite the outlined convergence of AI along neuromorphic principles, the research on neuromorphic AI has remained relatively separate from mainstream AI. The state of the art up to around 2022 is well summarized in a number of tutorial papers, including the November 2019 special issue of IEEE Signal Processing Magazine on ``Learning Algorithms and Signal Processing for Brain-Inspired Computing'' \cite{neftci2019surrogate,jang2019introduction,indiveri2019importance,rajendran2019low}, and  later reviews on event-driven sensing \cite{chen2020event} and neuroscience-inspired learning algorithms \cite{pehlevan2019neuroscience}. Other surveys have organized the field around algorithm-hardware co-design for spike-based machine intelligence \cite{roy2019towards}, and specific platforms such as Intel's Loihi processor \cite{davies2021advancing}. These reviews have established neuromorphic computing as a mature field with demonstrated applications in vision, robotics, and signal processing.

However, the last few years have seen an important shift in the design of neuromorphic AI platforms, which is highlighted in this article. As illustrated in Figure \ref{fig:tokenization}, modern AI models, including \textbf{transformers} and \textbf{state-space models} (SSMs), operate on sequences of \textbf{tokens} -- discrete informational units representing a (sub)word or an image patch as a vector consisting of multiple features or \textbf{channels}. Early work on neuromorphic AI has focused on the task of intra-token processing, i.e., of transforming individual tokens by combining different features, targeting applications such as image classification. In contrast,   recent research points to the potential of neuromorphic principles for sequence-to-sequence mapping via inter-token processing. This ongoing shift is the main focus of this article. 

\textbf{Intra-token processing} (Figure \ref{fig:intra}) transforms individual tokens independently by mixing information across channels within each token. This type of operation is exemplified by linear layers and, more broadly, conventional neural networks. Early neuromorphic systems, summarized in the previous reviews mentioned above, address this type of computation, focusing on tasks such as image classification. In these systems, the input data are represented and processed via sparse spiking (binary) signals defined along a \textbf{virtual time axis} -- an internal computational dimension  (see Figure \ref{fig:neuroAI}). 

Typical results, sketched in Figure \ref{fig:benefits} and exemplified in Table \ref{tab:intratoken_results}, show that, as computations are carried out along this virtual time axis, neuromorphic systems consume more energy, processing more and more spikes, eventually approaching  the energy consumption of a conventional system based on standard neural networks. The benefits of neuromorphic processing in this context materialize in a flexible trade-off between accuracy and energy: running the neuromorphic system for a shorter (virtual) time may cause a potentially small loss in accuracy but at much reduced energy consumption.

 \textbf{Inter-token processing}  (Figure \ref{fig:inter}) mixes information between different tokens in a sequence, separately for each channel. Notably, in transformers, \textbf{self-attention} mechanisms implement inter-token processing by computing pairwise relationships between all tokens; while in SSMs,  state-space dynamic updates carry information from past to future tokens. Recent work has explored the adoption of neuromorphic principles for the implementation of inter-token processing, encompassing both self-attention \cite{zhou2023spikformer,song2025xpikeformer,xing2024spikellm,pan2025spikingbrain} and SSMs \cite{stan2024learning,bal2024pspikessm,zhong2024spike}. In such systems, as illustrated in the bottom part of Figure \ref{fig:benefits} and exemplified in Table \ref{tab:intertoken_results}, improved energy-accuracy trade-offs arise from the sparsity of the inner representations leveraged by neuromorphic AI systems.

This article traces the evolution of neuromorphic systems from intra-token processing to inter-token processing, providing a unified introduction to this emerging topic that targets signal processing researchers.

\subsection{Organization}

This paper is organized as follows. Section \ref{sec:vs} establishes the mathematical foundations of token-based processing, formally defining intra-token and inter-token operations and explaining their role in modern AI architectures. Section \ref{sec:NPUmain} introduces \textbf{neuromorphic processing units}  (NPUs), a general model for neuromorphic AI systems, and its constituent blocks referred to as   \textbf{neuromorphic processing elements} (NPEs), formulating them as recurrent models with discrete activations.  Section \ref{sec:virtual} covers intra-token processing via computing along the virtual time axis. In contrast,  Section \ref{sec:inter1} and Section \ref{sec:inter2} discuss inter-token processing through the lens of associative memorization via state-space dynamics and self-attention, respectively. Section \ref{sec:train} reviews training methodologies, from teacher-forcing with surrogate gradients to biologically plausible local learning rules. Section \ref{sec:conclusions} concludes the paper.

A list of acronyms used in this article can be found in Table \ref{tab:acronyms}.

    \section{Intra-Token vs. Inter-Token Processing}\label{sec:vs}

\begin{figure}[!t]
\centering
\includegraphics[width=1\linewidth]{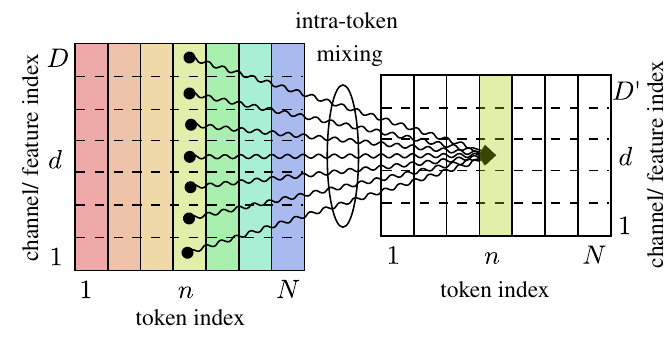}
\caption{Intra-token processing operates on each token $n$ independently, combining information across channels $d=1,...,D$ within a single token via a function of the form (\ref{eq:intra}). The output token has a generally different dimension $D'$.}
\label{fig:intra}
\end{figure}

\begin{figure}[!t]
\centering
\includegraphics[width=0.95\linewidth]{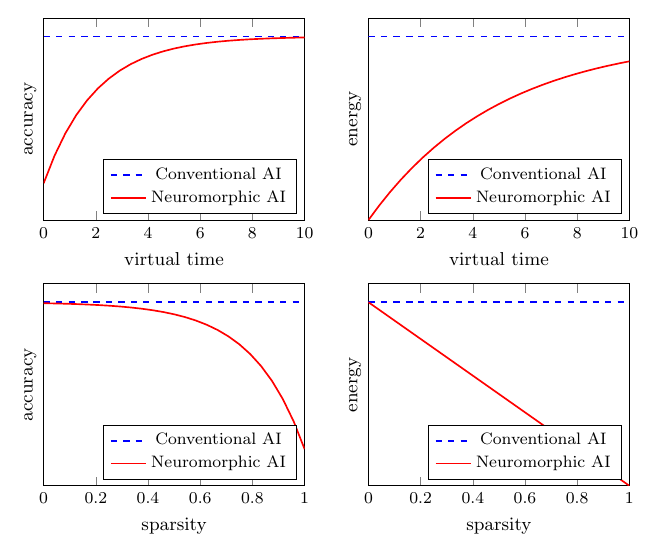}
\caption{Neuromorphic computing can enhance the energy efficiency of both intra-token processing and inter-token processing: (top) In most early work, the focus was on enhancing the energy efficiency of intra-token processing steps by operating over a virtual time axis via sparse spiking signals; (bottom) More recently, there has been interest in porting the benefits of dynamic sparsity via discrete activations, which is typical of neuromorphic processors, also to inter-token processing.}
\label{fig:benefits}
\end{figure}

Modern AI models, such as transformers, operate on data partitioned into \textbf{tokens}, which represent elementary units of information, like words in text or patches in images. Each token is encoded via a vector whose elements are referred to as \textbf{channels} or {features}. In this section, we first briefly review how token embeddings are created starting from data such as text or images. Then, we elaborate on the processing of tokens, which, as seen in Figure \ref{fig:arch}, typically  integrates  two complementary types of operations: {intra-token} and {inter-token} processing.

\subsection{Tokenization and Embedding}

Consider the case of an input text. As seen in Figure \ref{fig:tokenization}, the text is first partitioned into discrete word or subword units using algorithms like byte pair encoding  \cite{sennrich2016neural}, converting it into a sequence of token identifiers (IDs). This process is known as \textbf{tokenization}. Each token ID is an integer from the set of integers $\{1,...,V\}$, where $V$ is the size of the dictionary of possible tokens.

\begin{figure}[!t]
\centering
\includegraphics[width=0.6\linewidth]{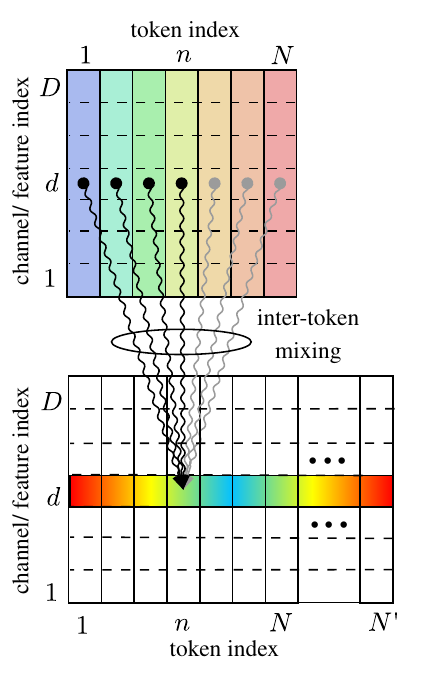}
\caption{Inter-token processing  operates on each channel $d$ independently,  mixing information from different tokens $n=1,...,N$ via a function of the form (\ref{eq:inter}). When masking is applied, only some input tokens can contribute to each $n$-th output token (the figure illustrates the case of an autoregressive mask, with inactive dependencies in gray). The number of output tokens $N'$ can exceed $N$, as is the case in generative language models (illustrated with suspension points). }
\label{fig:inter}
\end{figure}

Each token ID $v\in\{1,...,V\}$ is then mapped to a real-valued vector of dimension ${D_\text{emb}}$, which is read off a fixed dictionary of $V$ vectors. The vector assigned to each token ID is known as \textbf{token embedding}. Each entry of the embedding vector is known as a \textbf{channel} (or feature) of the embedding. Embeddings may also be pre-processed to include \textbf{positional} information about the index $n$, since the original embeddings alone do not carry this information. This is typically done by summing vectors or by multiplying by matrices that depend on index $n$ \cite{su2024roformer,dufter2022position}.

Overall, through the processes of tokenization and embedding, the  input text is converted into  a matrix $\bm{E} \in \mathbb{R}^{D_\text{emb} \times N}$, where  $N$ is the number of tokens and each $n$-th column of matrix $\bm{E}$, denoted as  $\bm{E}_{:,n}$, represents the $n$-th token embedding.  By construction, the token  index $n$ has the significance of time index, with time increasing as $n$ grows, starting from $n=1$. Note that it is also quite common to use the transpose of the matrix $\bm{E}$ as the result of the embedding step, so that tokens are represented as row vectors.

Similar transformations can also be applied to other types of inputs, including images and videos. To this end, an image is first divided into fixed-size patches, which are flattened and multiplied by a matrix to produce patch embeddings \cite{dosovitskiy2021image}. In this case, the token index $n$ does not represent a temporal index, but is rather an indicator for the  spatial position of the token in the image. Tokens are typically pre-processed to encoded this information.  For videos, frames are sampled temporally, and each frame is processed like an image, with additional temporal positional information added to capture the sequential nature of frames \cite{arnab2021vivit, bertasius2021spacetime}.

\begin{figure}[!t]
	\centering
	\includegraphics[width=0.8\linewidth]{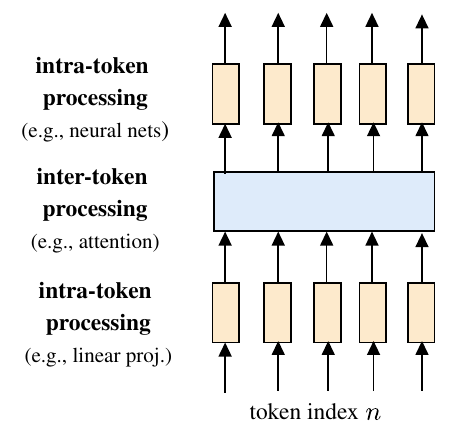}
	\caption{Modern AI systems typically alternate between intra-token processing steps, such as linear projections and feedforward neural networks (also known as multi-layer perceptrons), and inter-token processing, such as attention and state-based dynamics. The figure represents a single layer of computation. Note that  in this simplified picture, important blocks such as concatenation, normalization, multi-head processing, and residual connections are not represented.  }
	\label{fig:arch}
\end{figure}

\subsection{Modern AI Architectures}
As shown in Figure \ref{fig:arch}, modern AI architectures typically alternate between inter-token and intra-token processing steps. Starting from the token embeddings $\bm E$ (at the bottom of the architecture of Figure \ref{fig:arch}), tokens are transformed through a series of layers, with each layer involving both intra-token and inter-token operations. 

 \textbf{Inter-token} processing  allows tokens to communicate and gather contextual information from other positions in the sequence,  addressing the significance of each token in the \textbf{context} of other tokens of the sequence. This operation is carried out separately on each channel of the input tokens. 
 
 In contrast, \textbf{intra-token} processing  operates independently on each token to refine the features of the given token. When intra-token processing follows an inter-token processing step, the feature of a token are modified specifically  to account for the updated  representation produced by inter-token information exchange. Thus, the alternation between the two types of processing allows tokens to share information at the level of each channel, and then to individually process updated representations,   iteratively building  context-aware representations of the input data.

 To elaborate, write the input signal to a generic computational block  as the matrix $\bm X \in \mathbb{R}^{D\times N}$, where each column $\bm X_{:,n}$ is a \textbf{representation}  of the $n$-th token, with each representation being of dimension $D$. Depending on the model architecture, the dimension $D$ may or may not correspond to the initial dimension $D_\text{emb}$ of the token embeddings $\bm E$. Henceforth, we may refer to each token representation as a token for short.

In a similar manner,  denote the output sequence for the given computational block as $\bm{Y} \in \mathbb{R}^{D' \times N'}$, which includes the $N'$ output tokens $\bm Y_{:,n}$, each of dimension $D'$, as columns. The number of output tokens $N'$ and the number of output channels $D'$ may potentially differ from the input counterparts $N$ and $D$, respectively. The output tokens $\bm Y$ are also referred to as \textbf{activations}.

In the next two subsections, we describe how input tokens $\bm X$ are processed by intra-token and inter-token processing blocks to obtain the output tokens, or activations, $\bm Y$.

\begin{table*}[t]
\centering
\caption{Representative energy efficiency improvements at approximately equal accuracy levels for intra-token processing tasks on Intel Loihi neuromorphic hardware compared to GPU implementations with conventional neural networks }
\label{tab:intratoken_results}
\begin{tabular}{l c c}
\hline
\textbf{Task} & \textbf{Energy Efficiency Improvement} & \textbf{Reference} \\
\hline
Keyword spotting & $109\times$ & \cite[Figure~1]{blouw2019benchmarking} \\
CIFAR-10 classification & $100\times$ & \cite[Figure~4]{lenz2023ultra} \\
Image segmentation & $2\times$ & \cite[Figure~12]{patel2021spiking} \\
\hline
\end{tabular}
\end{table*}

\begin{table*}[t]
\centering
\caption{Representative Energy Efficiency Improvements for Inter-Token Processing Tasks Compared to Conventional Implementations (Theoretical Results Based on Operation Counts)}
\label{tab:intertoken_results}
\begin{tabular}{l l c l}
\hline
\textbf{Task} & \textbf{Architecture} & \textbf{Improvement} & \textbf{Reference} \\
\hline
Language modeling & RWKV-based SNN & $32.2\times$ & \cite[Table~1]{zhu2023spikegpt} \\
Language modeling & Self-attention (BERT-based) & $2\times$ & \cite[Figure~4]{bal2024spikingbert} \\
Long Range Arena & SSM-based SNN & $36\times$ & \cite[Sec.~4.2]{bal2024pspikessm} \\
Symbol demodulation & Self-attention (no masking) & $14 \times$ & \cite[Figure 8]{song2025xpikeformer}\\
\hline
\end{tabular}
\end{table*}

\subsection{{Intra-Token Processing}}

As illustrated in Figure \ref{fig:intra}, intra-token processing operates on each token independently, combining information across channels within a single token. Mathematically, given the  input $n$-th token  $\bm{X}_{:,n}$, each entry ${Y}_{d,n}$ of the corresponding $n$-th output token $\bm{Y}_{:,n}$ is produced via a transformation \begin{equation} \label{eq:intra} {Y}_{d,n} = f_{d,n}(\bm{X}_{:,n}),\end{equation} so that each channel $d$ of the $n$-th output token $\bm{Y}_{:,n} \in \mathbb{R}^{D'}$ depends only on the corresponding input token $\bm{X}_{:,n} \in \mathbb{R}^{D}$. As an example, a linear layer or a  feedforward network layer  implement this type of operation. 

By construction, the number of output tokens equals the number of input tokens, i.e., $N$, since each input token is mapped to a corresponding output token. In contrast, the output dimension $D'$, i.e., the number of channels,  may be different from the number of input channels $D$. 

\begin{table}[!t]
	\caption{List of Acronyms}
	\label{tab:acronyms}
	\centering
	\renewcommand{\arraystretch}{1.3}
	\begin{tabular}{p{0.18\columnwidth}p{0.72\columnwidth}}
		\hline
		\textbf{Acronym} & \textbf{Definition} \\
		\hline
		\hline
		AI & Artificial Intelligence \\
		BPTT & Backpropagation Through Time \\
		GPU & Graphical Processing Unit \\
		ID & Identifier \\
		LIF & Leaky Integrate-and-Fire \\
		NPE & Neuromorphic Processing Element \\
		NPU & Neuromorphic Processing Unit \\
		RWKV & Receptance Weighted Key Value \\
		SNN & Spiking Neural Network \\
		SSM & State-Space Model \\
		STDP & Spike-Timing-Dependent Plasticity \\
		XNOR & Exclusive NOR (logical operation) \\
		\hline
	\end{tabular}
\end{table}
    
\subsection{{Inter-Token Processing}} \label{sec:interfirst}As seen in Figure \ref{fig:inter},  inter-token processing  operates on each channel $d$ independently,  mixing information from different tokens $n=1,...,N$. Formally, given the input $d$-th features $\bm{X}_{d,:}$ across all tokens, the $d$-th feature of the $n$-th output token is given by the output of a function \begin{equation}\label{eq:inter} {Y}_{d,n} = g_{d,n}
    (\bm{X}_{d,:}). \end{equation}    Self-attention in transformers exemplifies this paradigm, computing token-to-token relationships independently for each channel. Note that, by construction, with inter-token processing, the number of output features, $D'$, equals the number of input features, i.e., $D'=D$.

  The function (\ref{eq:inter}) allows for an arbitrary dependence between output channel and the input channels across all tokens. This is important for applications such as image processing, in which each patch may benefit from information present in all other patches in the image. However,  in some cases it is useful, or even necessary,  to restrict the dependence of the $n$-th output token to a subset of the input tokens. For instance, using the \textbf{autoregressive mask} illustrated in Figure \ref{fig:inter}, the $n$-th token can only depend on the input tokens indexed $1,...,n$ (black arrows), implementing a \textbf{causal}  mapping with respect to the token index. 
   
   The autoregressive mask is most notably leveraged by \textbf{generative language models}. In such settings, the user \textbf{prompt} is encoded by the initial set of $N$ input token embeddings $\bm E$. These tokens are processed during a preliminary phase, often called \textbf{pre-fill}. The last output token, the $N$-th, is then \textbf{decoded} back into a token ID, and the corresponding embedding is added as a new column to matrix $\bm E$. The process is repeated for as many steps as needed, typically until some special ``stop'' token is generated. As a result, the number of output tokens, $N'$ is larger than the number of input tokens, $N$.

    \subsection{From Intra-Token to Inter-Token Processing}\label{sec:from}

The categorization into intra-token and inter-token processing provides a useful lens for understanding the evolution of the use cases and applications of neuromorphic AI models. This will be the main subject of this paper.

Standard feedforward neural networks are fundamentally models for intra-token processing, applying dense transformations to individual input vectors. Early applications of neuromorphic computing focused on the evaluation of \textbf{spiking neural networks} (SNNs) -- \textbf{recurrent}  neural networks with \textbf{binary}  activations (see next section) -- for \textbf{intra-token} operations, serving as energy-efficient replacements for conventional feedforward networks.  Corresponding tasks often focused on problems such as image classification, where each image constitutes a single token.

In such applications, the temporal dynamics of neuromorphic models is leveraged to carry out inter-channel mixing over a \textbf{virtual time} axis (see Figure \ref{fig:encoding}).  This may yield energy savings as compared to standard neural networks when the signals processed over this virtual time dimension are sparse and outputs can be produced in a few steps of the virtual time axis \cite{davies2021advancing} (see Figure \ref{fig:benefits}). In fact, energy consumption grows with the number of virtual time steps, while decreasing with the sparsity of the input.  Examples of representative results concerning the energy advantages of SNNs implemented over specialized hardware  against conventional neural networks run of GPUs can be found in Table \ref{tab:intratoken_results}.

Beyond intra-token mixing, the inherent recurrent, dynamic, nature of neuromorphic models  makes them suitable also for \textbf{inter-token} processing tasks. In fact, as detailed in the next section, recurrent models can naturally process a channel $d$ of a sequence of tokens by treating each token index $n$ as a discrete-time axis. Thus, unlike intra-token processing tasks, the discrete time axis underlying the operation of neuromorphic models for inter-token computing has the significance of  \textbf{real time} in the case of text data, and more broadly of positional information. 

Conventional examples of neuromorphic models processing information over real time encompass SNNs deployed to extract information from  event-driven data produced by neuromorphic sensors  \cite{schaefer2022aegnn,anumula2018feature,chen2023neuromorphic}. As discussed later in this article, more recent examples include spiking language models, in which discrete time runs over embeddings of a given text \cite{zhu2023spikegpt,yao2023spikedriven,xing2024spikellm,pan2025spikingbrain} and token-based vision models \cite{song2025xpikeformer}.  Table \ref{tab:intertoken_results} provides some examples of existing experimental findings  about the potential advantages of neuromorphic computing for tasks requiring inter-token processing.

	\section{Neuromorphic Processing Units }\label{sec:NPUmain}
	
	In this section, we introduce a general model for \textbf{neuromorphic  processing units} (NPU), a generalized architecture including SNNs as a special case.

\subsection{Defining Neuromorphic Processing Elements}\label{sec:charact}

\begin{figure*}[!t]
\centering
    \centering
    \includegraphics[width=0.8\linewidth]{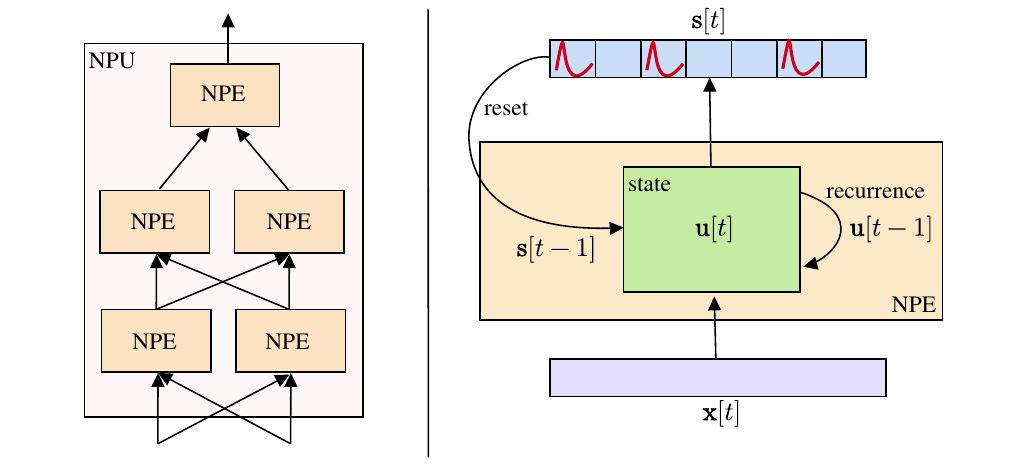}
    \caption{(left) A neuromorphic processing unit (NPU)  is constructed by composing multiple  {neuromorphic processing elements} (NPEs) in a network architecture, along with other blocks such as  concatenation, normalization, multi-head processing, and residual connections (not shown). (right) Each NPE operates over a discrete time index $t=1,2,...$, taking as input a vector $\bm{x}[t]$ and producing a vector output  $\bm{s}[t]$ at each time step $t$. Each NPE exhibits all -- or a subset -- of the following fundamental characteristics: discrete activations (i.e., spiking outputs), recurrence, and reset mechanisms.}
    \label{fig:NPU}
\end{figure*}

Neuromorphic models represent a bio-inspired paradigm for neural computation, where information is encoded and transmitted through discrete-time events, also known as \textbf{spikes}. As illustrated in the left panel of Figure~\ref{fig:NPU}, an NPU  is constructed by composing multiple \textbf{neuromorphic processing elements} (NPEs) in a network architecture. The network generally includes  other blocks such as concatenations, normalization, multi-head processing, and residual connections (not shown). 

As seen in the right panel of Figure~\ref{fig:NPU}, each NPE operates over a discrete time index $t$, taking as input a vector $\bm{x}[t]$ and producing a vector output  $\bm{s}[t]$ at each time step $t$. Specifically, in conventional SNNs, NPEs  take the form of individual \textbf{spiking neurons}, each receiving a vector $\bm x[t]$ as input through the neuron's synapses, and producing an a scalar output $s[t]$ at each discrete time $t$.  More generally, an NPE is a dynamic \textbf{multiple-input multiple-output} computational block, which exhibits all -- or a subset -- of the following fundamental characteristics:
\begin{itemize}
	\item \textbf{Discrete activations:} NPEs communicate through discrete activation values. These are most typically binary outputs, with 0 representing the absence of a spike and 1 representing a \textbf{spike} event. This discrete, event-driven communication stands in contrast to the continuous-valued activations of conventional artificial neural networks. This has two main benefits:\begin{itemize}
		\item \textbf{Dynamic sparsity:} Downstream NPEs can  avoid processing zero activations, since zero-valued inputs do not change their outputs. Therefore, a larger sparsity in the signals exchanged by NPEs -- in virtual or real time -- can yields important gains in terms of energy consumption, as long as the hardware makes it possible to adaptively gate unnecessary operations \cite{rathi2023exploring}. 
		\item \textbf{Multiplication-free operations:} When the activations are binary, multiplications with weight matrices  require only accumulate (sum)  operations, hence foregoing the need for significantly more complex multiply-and-accumulate operations. In fact, a multiplier typically consumes several times the energy of an adder, while also requiring an orders-of-magnitude larger chip area  \cite{horowitz20141}. 
	\end{itemize} 
	
	\item \textbf{Recurrence:} NPEs maintain internal states  that evolve over time, preserving information about past inputs and enabling temporal processing. For instance, the spiking neurons in an SNN maintain a membrane potential  that is updated at each time step as a function of the current input and previous state. This recurrent dynamics allows NPEs to integrate information over time and respond to temporal patterns in the input stream, making them naturally suited for sequence processing tasks.
	
	\item \textbf{Reset mechanism:} The internal state of an NPE is influenced not only by external inputs but also by the NPE's own outputs, often through a \textbf{reset mechanism}. In particular, when an NPE produces a non-zero output, emitting a spike, its internal state undergoes a reset operation that returns it to a baseline level. This reset creates a nonlinear feedback loop that may help stabilize the NPE by controlling   the neuron's firing patterns and information encoding properties. 
\end{itemize}

	\subsection{Modeling Neuromorphic Processing Elements }\label{sec:subrec}

	As illustrated in the right panel of Figure  \ref{fig:NPU},  consider an NPE processing an input sequence $\{\bm{x}[t]\}_{t=1}^T$ where $\bm{x}[t] \in \mathbb{R}^{D}$ represents the input at discrete time step $t$.  The time index $t$  has  different interpretations depending on whether the NPE implements intra-token or inter-token processing. In particular, as anticipated in Section \ref{sec:from}, the time index $t$ has the significance of \textbf{virtual time} in the case of intra-token processing, while for inter-token processing it represents a variable running over \textbf{real time} or, more generally, over positional information.

    \subsubsection{\textbf{Discrete activations}}  An NPE maintains an internal state vector $\textbf{u}[t]=[u_1[t],...,u_{D_\text{h}}[t]]^\top \in \mathbb{R}^{D_\text{h}}$ as a function of its inputs over time $t$. The subscript ``$\text{h}$'' emphasizes that the internal state vector is hidden, or latent. For conventional spiking neurons in SNNs, the internal state is a scalar known as the \textbf{membrane potential}.  The output generation process is governed by a threshold mechanism, and it produces a discrete $D_\text{h}$-dimensional signal $\textbf{s}[t]$, with the $i$-th output $s_i[t]$ being a function only of the corresponding $i$-th state entry $u_i[t]$. 
    
    In NPEs, outputs, i.e., activations, are ideally \textbf{sparse}, in the sense that a small number of the outputs $s_i[t]$ is non-zero at any given time. In general, one says that the output corresponds to a \textbf{spike} when it is non-zero, i.e., 
\begin{equation}
s_{i}[t]\begin{cases}
=0 & \text{\text{no spike}}\\
\neq 0 & \text{spike}.
\end{cases}
\end{equation}A spike is represented as an impulsive waveform in the representation of the activations given in the right panel of Figure \ref{fig:NPU} (and a stem in Figure \ref{fig:encoding}). 

In this regard, one can distinguish the following types of output mechanisms:
    \begin{itemize}
        \item \textbf{Deterministic binary spikes:} In the most common case of binary outputs, for each output $i$, a spike is emitted deterministically when the membrane potential exceeds a threshold $\gamma_{\text{th}}$, as 
	\begin{align}
s_{i}[t] & =\Theta(u_{i}[t]-\gamma_{\text{th}})\nonumber \\
 & =\begin{cases}
0\text{ (no spike)} & \text{if }u_{i}[t]\leq\gamma_{\text{th}}\\
1\text{ (spike)} & \text{if }u_{i}[t]>\gamma_{\text{th}},
\end{cases}\label{eq:spike_generation}
\end{align}
	where $\Theta(\cdot)$ is the Heaviside step function. In this case, each spike carries no {payload} in the sense that information is only conveyed in the timing $t$ of the spike. 
      \item \textbf{Probabilistic binary spikes:} 
    A probabilistic version of the binary spiking mechanism (\ref{eq:spike_generation}) draws the output from a  Bernoulli distribution with spiking probability given by $\Pr[s_{i}[t]=1|u_{i}[t]]=\sigma(u_{i}[t]-\gamma_{\text{th}})$, where   $\sigma(\cdot)=(1+\exp(-\cdot))^{-1}$ is the sigmoid function, i.e.,  \begin{equation}\label{eq:prob} s_i[t] \sim \text{Bern}(\sigma(u_i[t] - \gamma_\text{th})).\end{equation} This way, the spiking probability increases as the state variable $u_i[t]$ grows beyond the threshold $\gamma_\text{th}$.
    \item \textbf{Signed spikes:} For ternary, or signed, spikes ($L=3$), the spike output takes values in $\{-1, 0, +1\}$, enabling both ``excitatory'' (positive) and ``inhibitory'' (negative) spike emissions. The spike generation rule depends on two  thresholds $\gamma^1_{\text{th}}$ and  $\gamma^2_{\text{th}}$, with $\gamma^1_{\text{th}}<\gamma^2_{\text{th}}$, as
	\begin{equation}
		s_i[t] = \begin{cases}
			-1, & \text{if } u_i[t] \leq  \gamma_{\text{th}}^1, \\
			+1, & \text{if } u_i[t] > \gamma_{\text{th}}^2, \\
			0, & \text{otherwise},
		\end{cases}
		\label{eq:ternary_spike}
	\end{equation}
    \item \textbf{Multi-level spikes: }  Generalizing the binary mechanism (\ref{eq:spike_generation}), the state $u_i[t]$ is quantized into $L+1$ discrete levels (including 0)  through multiple thresholds $\{\gamma_{\text{th}}^k\}_{k=1}^{L}$ as
	\begin{equation}
		s_i[t] = \sum_{k=1}^{L} \Theta(u_i[t] - \gamma^k_\text{th}),
		\label{eq:gif_spike}
	\end{equation}
	where $\gamma_\text{th}^1 < \gamma_\text{th}^2 < \cdots < \gamma_\text{th}^{L}$.  The output (\ref{eq:gif_spike}) indicates the absence of a spike if $u_i[t] \leq \gamma^1_\text{th}$. Otherwise, it  corresponds to an integer $s_i[t]\in\{1,...,L\}$, which reports the quantization level for the input $u_i[t]$. Thus, information-theoretically, each multi-level spike carries information  not only through its time $t$, but also through its playload, which can encode up to additional $\log_2(L)$ bits.
    \end{itemize}

\subsubsection{\textbf{Internal State}}	 As discussed, each NPE maintains an internal state  $\bm{u}[t] \in \mathbb{R}^{D_\text{h}}$. The state  evolves according to first-order  discrete-time dynamics
	\begin{equation}
		\bm{u}[t] = f(\bm{u}[t-1], \bm{x}[t], \bm{s}[t-1]; \bm{\theta}),
		\label{eq:general_dynamics}
	\end{equation}
	where  $\bm{\theta}$ represents  learnable parameters. This general model can capture  recurrence through its dependence on the past state vector $\bm{u}[t-1]$, as well as  the reset mechanism through the past outputs $\bm{s}[t-1]$. 

     The most widely adopted dynamics model is the \textbf{leaky integrate-and-fire} (LIF) mechanism, which implements an exponential decay of the membrane potential coupled with input integration. The corresponding dynamics (\ref{eq:general_dynamics}) are defined as 
	\begin{equation}
		\bm{u}[t] = \alpha \bm{u}[t-1] + \bm{W}_{\text{in}}\bm{x}[t] -  \bm{W}_{\text{res}}\bm{s}[t-1], 
		\label{eq:lif_dynamics}
	\end{equation}
	where $\alpha \in (0,1]$ is a state  decay factor, $\bm{W}_{\text{in}} \in \mathbb{R}^{D_\text{h} \times D}$ denotes the input weight matrix, and $\bm{W}_{\text{res}} \in \mathbb{R}^{D_\text{h} \times D_\text{h}}$ represents feedback  connection from output to state. 

    As discussed in the next sections, the input weight matrix $\bm W_\text{in}$ plays different roles for intra- and inter-token processing. For instance, in the case of conventional spiking neurons in an SNN, the entries of matrix $\bm W_\text{in}$ describe the synaptic weights applied by the constituent spiking neurons to combine their inputs $\bm x[t]$ and evaluate their updated membrane potentials $u_i[t]$. 
    
    The feedback matrix $\bm W_\text{res}$  can be instead used to implement the reset mechanism. Take for instance the case of binary spikes (\ref{eq:spike_generation}). When a spike is produced, i.e., when $s_i[t]=1$, a reset operation decreases the corresponding state $u_i[t]$,  which just crossed the threshold $\gamma_\text{th}$, by an amount equal to  the threshold $\gamma_\text{th}$. In order to implement this mechanism using the dynamics (\ref{eq:lif_dynamics}), one selects the feedback matrix as a diagonal matrix with diagonal elements all equal to the threshold $\gamma_\text{th}$, i.e., \begin{equation}\label{eq:res}\bm{W}_{\text{res}}=\gamma_\text{th} \bm I,\end{equation} where $\bm I$ is the identity matrix. Similar transformations also apply in the case of ternary and multi-level spikes.

	The  LIF model maintains a real-valued state vector $\bm u[t]$, which is updated via a scalar gain $\alpha$. Alternatively, there exist spiking neural models that implement higher-order  dynamics by introducing more than one state variable per output $y_i[t]$ and by applying matrix gains in the state update. This approach  enables richer temporal dynamics, including oscillatory behavior. Accordingly, these models are particularly well suited to model periodic patterns and, more broadly, signals with informative spectral features, such as audio or radio signals \cite{izhikevich2001resonate,higuchi2024balanced,wu2025neuromorphic}.   As an alternative to state-based dynamics of the form (\ref{eq:lif_dynamics}),  transition models based on look-up tables operating at the level of groups of timings  have also been   proposed    \cite{izhikevich2025spikingmanifesto}.

    \section{Neuromorphic Processing Units for Intra-Token Processing}\label{sec:virtual}
As discussed in Section \ref{sec:from}, earlier work on NPUs has focused on the use of SNNs for intra-token processing tasks, such as linear projections (i.e., matrix-vector multiplications) and, more generally, feedforward neural networks. This section reviews the operation of NPUs for intra-token processing by focusing on methods that have found applications in modern AI models. 

\subsection{Input Encoding Along the  Virtual Time Axis}

    \begin{figure*}[!t]
\centering
    \centering
    \includegraphics[width=0.7\linewidth]{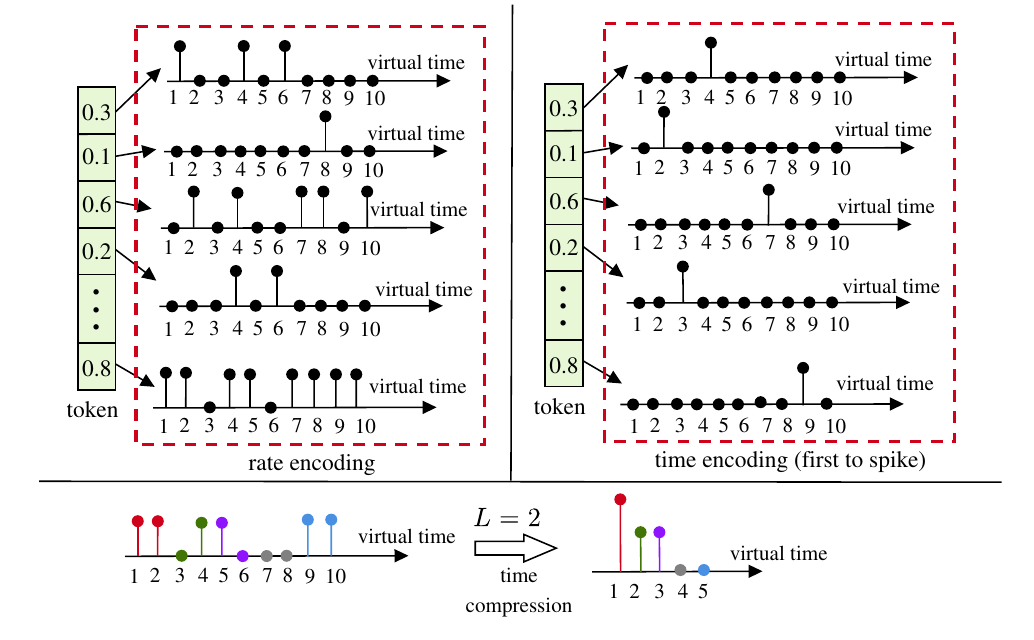}
    \caption{(top-left) With rate encoding, the entries of a token are encoded one by one into a binary  signal with information encoded in the number of spikes over the virtual time axis. (top-right) With time encoding, the entries of a token are encoded one by one into a binary  signal with information encoded in the timing of the spikes over the virtual time axis. (bottom) Using multi-level spikes allows the compression of the virtual time axis -- the figure illustrates the case with rate encoding.}
    \label{fig:encoding}
\end{figure*}

    In the case of {intra-token} processing, each input token $\bm X_{:,n}$, a $D$-dimensional vector,  is processed separately. Accordingly, in the rest of this section we will not always make the dependence on the token index $n$ explicit, since the discussion applies equally for any fixed value token index $n$. 
    
    Given a real-valued input token $\bm X_{:,n}$, the first step in the deployment of an NPU  is the  encoding of the vector  $\bm X_{:,n}$ as a signal that evolves along the virtual time axis $t$. This conversion  is typically  necessary at the input of the overall NPU, i.e., at the bottom of the architecture in Figure \ref{fig:arch}.   However,  in some applications,  the input to the NPU is already in the form of  spiking signals. This is the case for NPUs   that process  inputs from neuromorphic sensors, such as as event-driven cameras, that produce  spiking signals \cite{chen2020event}.

    Through this preliminary conversion step, the $D$-dimensional real-valued  token representation  $\bm X_{:,n}$  is  converted into a time sequence $\bm{x}[1],\bm{x}[2],...,\bm{x}[T]$ of $D$-dimensional discrete-valued vectors $\bm x[t]$ for $t=1,...,T$. The vectors $\bm x[t]$ are typically binary, taking values in the set $\{0,1\}$ but they may also be ternary (taking values in set $\{-1,0,+1\}$) or multi-level (taking values in set $\{0,1,...,L\}$). Henceforth, we will refer to any signal with discrete entries as a \textbf{spiking signal}, with the understanding that such signals are ideally \textbf{sparse}, i.e., having mostly zero entries.

    The time index $t=1,...,T$ marks  time steps in a  {virtual time} axis, with $T$ being the number of time steps of the token encoding. Importantly, the time index $t$ runs from 1 to $T$ separately for each token, thus decoupling the token index $n$, which may represent real time, from the virtual time $t$. The function of the virtual time $t$ is purely computational, describing the duration of the process of producing output token $\bm Y_{:,n}$ from the input token $\bm X_{:,n}$.

    The conversion between the token representation  $\bm X_{:,n}$ and the sequence $\bm{x}[1],\bm{x}[2],...,\bm{x}[T]$ along the virtual time axis can take place using different encoding methods. In all typical cases reviewed below, as illustrated in Figure \ref{fig:encoding}, each $d$-th token entry $X_{d,n}$ is encoded into a separate binary time series $x_d[1],...,x_d[T]$. Accordingly, we have the equality  $D_\text{h}=D$ between the dimensionalities of the input token representation $\bm X_{:,n}$ and of the NPE's inputs. 

    \subsubsection{\textbf{Rate Encoding}}  The most common virtual time-axis encoding strategy is \textbf{rate encoding}. As  seen in Figure \ref{fig:encoding} (top-left panel), rate encoding  produces a time series such that  the average value of the input $ x_d[t]$ over virtual time $t$ represents the original value $X_{d,n}$, i.e., \begin{equation}\label{eq:rateenc}
		X_{d,n} =   \frac{1}{T}\sum_{t=1}^T x_d[t].
	\end{equation} Rate encoding (\ref{eq:rateenc}) can represent numbers $X_{d,n} $ in the interval $[0,1]$, but scaling and and shifting allow the representation of any bounded quantity. As further discussed below, the resolution of this representation  depends on the duration $T$.
    
    With rate  encoding (\ref{eq:rateenc}) the positions of the ones (spikes) in the sequence have no bearing on the output, and they  can be determined in different ways. Options include placing the spikes in the $X_{d,n}T$ lowest indices $t$ and selecting $X_{d,n}T$ random time instants in the set $\{1,...,T\}$. In the first case, rate encoding is also known as  {unary coding}. 
    
  \subsubsection{\textbf{Bernoulli Encoding}}  Bernoulli encoding  is a stochastic variant of rate encoding. In Bernoulli encoding, the sequence $x_d[1],...,x_d[T]$ is generated i.i.d. from a Bernoulli distribution with probability $X_{d,n} \in [0,1]$, i.e., \begin{equation}\label{eq:Bernoulli}x_d[1],...,x_d[T]\underset{\text{i.i.d.}}{\sim} \text{Bern}(X_{d,n}).
    \end{equation} This way, the condition (\ref{eq:rateenc}) is approximately satisfied as $T$ increases.

    \subsubsection{\textbf{Time Encoding}} Rate encoding is information-theoretically inefficient, as it encodes $\log_2(T+1)$ bits in $T$ virtual time steps. In fact, there are only $T+1$ values that the rate (\ref{eq:rateenc}) can take. Other, potentially more efficient, encoding strategies include {time encoding}, in which the different virtual times $t \in \{1,...,T\}$ play a distinct role. In particular, with \textbf{positional time  encoding}, the $T$ bits $x_d[1],...,x_d[T]$ represent a $T$-bit integer, thus carrying up to $T$ bits of information. More constrained forms of time encoding, e.g., based on first-to-spike time encoding or relative-timing encoding, have a generally lower capacity \cite{jang2019introduction,izhikevich2025spikingmanifesto} (see the top-right panel of Figure \ref{fig:encoding} for an illustration of first-to-spike encoding).

 \subsubsection{\textbf{Multi-level spikes}} The virtual time horizon $T$ can be potentially compressed by using multi-level spikes instead of binary spikes. For example, assuming rate encoding,  each $L$-ary spike  can encode the information contained in $L$ virtual time steps \cite{xing2024spikellm} (see Figure \ref{fig:encoding} for $L=2$). As another example, with positional time encoding, each $L$-ary spike can encode $\log_2(L+1)$ time steps.

 \subsubsection{\textbf{Decoding from Virtual Time Axis to Reals}}  The spiking output $\bm{s}[1],...,\bm{s}[T]$ of the NPU (see, e.g., the top NPE in Figure \ref{fig:NPU}) can be converted to a real-valued output $\bm Y_{:,n}$ using the inverse mapping corresponding to rate or time encoding. For instance, with rate decoding, the output $Y_{d,n}$ is the average spiking rate of the corresponding output signals ${s}_d[1],...,{s}_d[T]$, i.e., $Y_{d,n}=(\sum_{t=1}^T s_d[t])/T$.

 \subsection{Processing Along the Virtual Time Axis}

As discussed in this section, once the input is converted into a time series $\bm x[1],...,\bm x[T]$ along the virtual time axis, it can be directly fed as input to a NPU consisting of a network of NPEs as illustrated in Figure \ref{fig:arch}. 
    
Most applications of NPUs to intra-token processing apply standard \textbf{multi-layer SNNs}. A multi-layer SNN consists of the concatenation of multiple layers, with each layer   implementing LIF-based dynamics  (\ref{eq:lif_dynamics}) and communicating to the next-layer NPE via binary (\ref{eq:spike_generation}), ternary \eqref{eq:ternary_spike}, or multi-level (\ref{eq:gif_spike}) discrete activations. Specifically, each layer consists of $D_\text{h}$ separate \textbf{spiking neurons}, where each neuron $i$ maintains the corresponding state variables $u_i[t]$ in vector $\bm u[t]$. This corresponds to a special case of the LIF-based dynamics (\ref{eq:lif_dynamics}) with diagonal matrix $\bm W_\text{res}$. The number of spiking neurons $D_\text{h}$ generally varies across layers.

In this standard  setting, each $i$-th spiking neuron in any layer is a \textbf{multiple-input single-output} block, which process the entire input $\bm x[t]$ from the previous layer to  produce the scalar spiking output $s_i[t]$. By the LIF-based dynamics (\ref{eq:lif_dynamics}),  this is done by first updating the state variable $u_i[t]$ as \begin{equation} u_i[t]
= \alpha u_i[t]+ \bm w_i \bm x[t],\end{equation} where $\bm w_i$ is the vector synaptic weights obtained as the  $i$-th row of matrix $\bm W_\text{in}$; and then evaluating the discrete activation $s_i[t] $ via (\ref{eq:spike_generation}), \eqref{eq:ternary_spike}, or (\ref{eq:gif_spike}).
  

	\section{Neuromorphic Processing Units for Inter-Token Processing via State-Space Dynamics}\label{sec:inter1}

    The previous section has discussed the conventional task of intra-token processing, and we now turn to more recent advances on inter-token processing via NPUs.  There are currently two main paradigms for inter-token processing: state-space dynamics, adopted by SSMs, and self-attention, which is adopted by transformers. This section addresses state-space dynamics, reviewing the implementation of SSMs via NPUs.  We start by presenting the general framework of inter-token mixing via associative memorization, which underlies both state-space dynamics and self-attention.

\subsection{Inter-Token Mixing as Associative Memorization} 

The intra-token mixing steps applied by both SSMs and transformers can be viewed as forms of \textbf{associative memorization}. Associative memories implement \textbf{content-addressable} information retrieval mechanisms. Unlike conventional memories,  the output is obtained not based on a specific address, but  by matching  cue to stored patterns based on content similarity \cite{krotov2025modern}. 
 
Associative memories rely on three main quantities:
\begin{itemize}
	\item \textbf{Value} $\bm{v}$: The value vector $\bm v \in \mathbb{R}^{D_\text{v}}$ represents the actual content or information to be stored and later retrieved. 
	
    \item \textbf{Key} $\bm{k}$: The key vector $\bm k \in \mathbb{R}^{D_\text{k}}$ plays the role of a continuous-valued vector index associated with a value vector $\bm{v}$. In memory terms, the key serves as an identifier for the content being stored. Importantly, this identifier does not take the form of a hard index, but rather as a soft representation of the content associated with vector $\bm v$.

    \item \textbf{Query} $\bm{q}$: The query vector $\bm q \in \mathbb{R}^{D_\text{k}}$  represents a new cue used to retrieve information from memory. The query is matched against stored keys to determine which values should be recalled and with what strength. Note that query and key vectors share the same dimension $D_\text{k}$.
\end{itemize}

In an inter-token processing system,  a triple of value, key, and query vectors are produced for each input token. This is done via preliminary intra-token operations, typically linear projections.  To elaborate, denote as $\bm Z \in \mathbb{R}^{D_\text{z}\times N}$ the  matrix of $N$ token representations produced by  the previous computational block in the given model architecture. The dimension $D_\text{z}$ typically coincides with the embedding dimension $D_\text{emb}$.

Value, key, and query vectors are computed via \textbf{linear projections} of each input token representation $\bm Z_{:,n}$ as \begin{equation}\label{eq:qkv} \bm V_{:,n}=\bm W_\text{v} \bm Z_{:,n}, \text{ } \bm K_{:,n}=\bm W_\text{k} \bm Z_{:,n},  \text{ and } \bm Q_{:,n}=\bm W_\text{q} \bm Z_{:,n},\end{equation}  with matrices $\bm W_\text{v} \in \mathbb{R}^{D_{\text{v}} \times D_{\text{z}}}$, $\bm W_\text{k} \in \mathbb{R}^{D_{\text{k}} \times D_\text{z}}$, and $\bm W_\text{q} \in \mathbb{R}^{D_{\text{k}} \times D_\text{z}}$. Equivalently, these transformations can be expressed in terms of the value matrix $\bm V \in \mathbb{R}^{D_\text{v}\times N}$, key matrix $\bm V \in \mathbb{R}^{D_\text{k}\times N}$, and query matrix $\bm V \in \mathbb{R}^{D_\text{k}\times N}$ as \begin{equation}\label{eq:qkv1} \bm V=\bm W_\text{v} \bm Z, \text{ } \bm K=\bm W_\text{k} \bm Z,  \text{ and } \bm Q=\bm W_\text{q} \bm Z.\end{equation}  The linear projections in (\ref{eq:qkv}) and (\ref{eq:qkv1}) are intra-token processing steps, and hence they can be evaluated using NPEs that apply the methods discussed in the previous section.

Inter-token processing implements an operation of the form  (\ref{eq:inter}) at the level of the value vectors. Accordingly, the value entry $ V_{n,d}$ plays the role of the input $ X_{n,d}$ in (\ref{eq:inter}). The rest of this section describes a notable instantiation of the function (\ref{eq:inter}) via state-space dynamics, while the next section focuses on alternative implementation based on self-attention.

As a side note, it is also possible to formalize intra-token processing in two-layer feedforward neural networks, which are typically adopted in transformer architectures, as a form of associative memory. However, unlike the dynamic and adaptive  associative memory mechanism applied by inter-token mixing,  intra-token processing effectively implements a static memorization mechanism \cite{geva2021transformer}.
	
	\subsection{Inter-Token Processing via State-Space Models}\label{sec:SSMfund}
	
	Inter-token processing via state-space dynamics processes the value, key, and query vectors (\ref{eq:qkv}) over time $t=1,2,...$. Specifically, the $t$-th triple $(\bm V_{:,t},\bm K_{:,t},\bm Q_{:,t})$ is processed at time $t$. The token index $n$ hence coincides with the computational time $t$, i.e., $n=t$.  This is in contrast to intra-token processing, which relies on computing over a virtual time axis, thus decoupling token index $n$ and computing time $t$. 
	
	Given the highlighted mapping between token index $n$ and computing time $t$, henceforth in this section we write the sequence of values, keys, and queries as $\bm v[t]=\bm V_{:,t}$, $\bm k[t]=\bm K_{:,t}$,, and $\bm q[t]=\bm Q_{:,t}$, respectively, for $t=1,2,...$

	Given the sequence of value, key, and query vectors, inter-token mixing via state-space dynamics operates separately on each channel $v_d[t]$ of the value sequence, recombining past values $\{v_d[t']\}_{t'=1}^t$ with weights increasing with the alignment of the corresponding key vectors $\{\bm k[t']\}_{t'=1}^t$ for the current query vector $\bm q[t]$.

	    To this end, for each $d$-th channel, SSMs maintain a vector state $\bm h_d[t] \in \mathbb{R}^{D_\text{k}}$ of the same dimensions, $D_\text{k}$,  of the key and query vectors as \cite{lahoti2025chimera}\begin{equation} 	\bm{h}_d[t] = a[t]\bm{h}_d[t-1] + b[t] \bm k[t] v_d[t], \label{eq:ssm_discrete}\end{equation} where the scalar gains $a[t]$ and $b[t]$ can be chosen as fixed constants, i.e., $a[t]=a$ and $b[t]=b$, or as function of the original input token $\bm Z_{:,t}$, as further discussed below. By (\ref{eq:ssm_discrete}), the state $\bm h_d[t]$ memorizes the association between the $d$-th channel of the value, $v_d[t]$, and the corresponding key $\bm k[t]$. 

    The $d$-th channel $ Y_{d,t}$ of the output token $\bm Y_{:,t}$, denoted as $y_d[t]$,  is then obtained by projecting the state $\bm h_d[t]$ into the current key vector $\bm q[t]$ as   \begin{equation}
		y_d[t] = \bm{q}[t]^\top\bm{h}_d[t].
		\label{eq:ssm_discrete_output} 
    \end{equation} This operation has the effect of giving more weight to values $v_d[t']$ whose keys $\bm k[t']$ are more aligned with the query $\bm q[t]$.

    To see this, consider the choice $a[t]=1$ and $b[t]=1$, which yields the \textbf{linear attention model} as a special case of the model (\ref{eq:ssm_discrete})-(\ref{eq:ssm_discrete_output}). With this choice, the inner state (\ref{eq:ssm_discrete}) is given by  \begin{equation} 	\bm{h}_d[t] = \sum_{t'=1}^t  \bm k[t'] v_d[t'], \label{eq:ssm_sum_la}\end{equation} and thus the output token has entries \begin{equation} 	y_d[t]= \sum_{t'=1}^t  S_{t,t'} v_d[t'], \label{eq:ssm_sum_la1}\end{equation} with coefficients \begin{equation}\label{eq:linearattfirst} S_{t,t'}=\bm{q}[t]^{\top}\bm{k}[t']. \end{equation}  The output (\ref{eq:ssm_sum_la1}) corresponds to a linear combination of the past values,  with weights given by the inner products between the corresponding query and key values. 
    
    The coefficient $S_{t,t'}$, measuring the similarity between query $\bm q[t]$ and key $\bm k[t']$ is also known as an \textbf{attention score}. In fact, the coefficient $S_{t,t'}$ indicates how much ``attention'' the $t$-th token should pay to the $t'$-th token in the linear combination (\ref{eq:ssm_sum_la1}). This concept will be further expanded on in the next section when discussing self-attention.

    Beyond linear attention, which adopts constant scalar gains $a[t]$ and $b[t]$, SSMs have evolved along two important directions:\begin{itemize}
    	\item \textbf{Gating:} The gains $a[t]$ and $b[t]$ in the state-space dynamics (\ref{eq:ssm_discrete}) can be adapted over time $t$ to  implement  \textbf{gating} mechanisms. Gating  modulates the extent to which input $\bm k[t]v_d[t] $ is memorized via the state $\bm h[t]$. This yields the class of \textbf{selective SSMs}. A common choice is $a[t]=\exp(-\Delta[t])$ and $b[t]=\Delta[t]$, where $\Delta[t]$ is a non-negative function of the input $\bm Z_{:,t}$ \cite{lahoti2025chimera}.  This way,   a larger value of the time step  $\Delta[t]$  causes the value channel $v_d[t]$ to affect the state $\bm h_d[t]$ to a larger extent, partially ``forgetting'' past values. A similar mechanism can be potentially implemented also for intra-channel mixing \cite{behrouz2024mambamixer}.
    	\item \textbf{Matrix-based gains:} Instead of using a scalar gain $a$, it is possible to multiply the past value vector $\bm h[t-1]$ with  a matrix $\bm A$. This matrix can be designed to approximate dynamic projections into an evolving orthogonal basis \cite{gu2022hippo}.  
    \end{itemize}

    
    

     \subsection{NPEs for Inter-Token Processing via State-Space Dynamics}\label{sec:real}

 In the special case of constant gains $a[t]=a$ and $b[t]=b$, the dynamics (\ref{eq:ssm_discrete}) of an SSM can be implemented by a LIF-based NPE (\ref{eq:lif_dynamics}) by mapping the SSM state $\bm h_d[t]$ to the NPE state $\bm u[t]$, the gain $a$ to the decay coefficient $\alpha$, the input $\bm x[t]$ to the key-value product $\bm k[t] v_d[t]$, the gain $b$ to the diagonal matrix $\bm W_\text{in}= b \bm I$, and by removing the reset term in (\ref{eq:lif_dynamics}), i.e., by setting $\bm W_\text{res}=\bm 0$. It is emphasized again that, unlike the case of intra-token processing discussed in the previous section,  here the computing  index $t$ coincides with the token index $n$.

Beyond the presence of the reset mechanism, the LIF-based NPE, however,  deviates from an SSM (\ref{eq:ssm_discrete})-(\ref{eq:ssm_discrete_output}) due to the adoption of  non-linear activation functions (\ref{eq:spike_generation})-(\ref{eq:gif_spike}). Stacking NPEs with reset-free dynamics and with the probabilistic spiking mechanism (\ref{eq:prob}) yields the \textbf{P-SpikeSSM} architecture introduced in~\cite{bal2024pspikessm} (see also \cite{stan2024learning} for a similar architecture). Neuromorphic models that retain the reset mechanism were studied in \cite{du2024spikingssms,zhong2024spike}. There are also spiking SSMs that implement linear projections via intra-token NPEs along the virtual time axis \cite{huang2025spikingmamba}.

\subsection{RWKV}
Inter-token processing based on state-space dynamics of the form (\ref{eq:ssm_discrete}) leverages a  state vector to memorize past key-value associations. This memory is then used to retrieve the new token value by projecting the state onto the new query vector. In contrast, the \textbf{receptance weighted key value model} (RWKV) implements a form of inter-token processing by  directly mixing value vectors element-wise before producing the output via a gating mechanism followed by a linear projection~\cite{peng2023rwkv}. Value mixing takes place in a recurrent way, implementing a form of state-space dynamics \cite[Appendix D]{peng2023rwkv}.

%

 \begin{figure}[!t]
\centering
\includegraphics[width=0.7\linewidth]{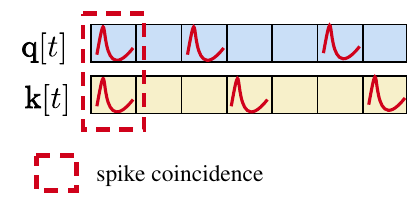}
\caption{Spiking transformer models evaluate inner products (here between query and key values) by spike coincidence counts (using AND gates) or binary digits coincidence counts (using XNOR gates). In this example, the former approach yields a score of 1, since there is a single spike coincidence, while the latter yields a score of 3, since the two sequences coincide in three positions. }
\label{fig:attentionsp}
\end{figure}


In a manner similar to the relationship between SSMs and P-SpikeSSM,	\textbf{SpikeGPT} adapts RWKV for neuromorphic computing by following the described RWKV processing with a layer of LIF neurons producing binary activations \cite{zhu2023spikegpt}.

\section{Neuromorphic Processing Units for Inter-Token Processing via Self-Attention}\label{sec:inter2}

As seen in the previous section, inter-token mixing can be designed based on associative memory mechanisms. The previous section has covered an implementation of such mechanisms based on state-space dynamics, which can be naturally realized by NPEs. This section discusses an alternative implementation based on the self-attention approach used by transformers.

\subsection{Softmax Self-Attention}

The linear attention model (\ref{eq:ssm_sum_la1})-(\ref{eq:linearattfirst}) effectively summarizes the entire history of key-value associations for the $d$-th channel in the state vector $\bm h_d[t]$ in (\ref{eq:ssm_discrete}). This yields an inter-token processing operation that has two potentially important limitations: \begin{itemize}
    \item \textbf{Insufficient selectivity:} The linear attention weights $S_{t,t'}$ in (\ref{eq:linearattfirst}) may turn out to be approximately equal for many values $\bm v[t']$, thus potentially lacking sufficient selectivity in the information retrieval process.
    \item \textbf{Autoregressive-only generation:} The mechanism (\ref{eq:ssm_sum_la1})-(\ref{eq:linearattfirst}) inherently applies an autoregressive mask, making the $t$-th output a function only of the previous inputs $t'\in\{1,...,t\}$. As discussed in  Section \ref{sec:interfirst}, while this type of inter-token processing is important in applications such as language models, other tasks require bidirectional attention schemes in which each token can attend to every other token. 
\end{itemize} Softmax self-attention addresses these problems by introducing a different inter-token mixing mechanism whereby (\emph{i}) the query-key inner products are pre-processed to magnify the largest inner products against the rest, and (\emph{ii}) each current token $n$ attends directly and separately to all tokens $n'\in \{1,...,N\}$, and not just to previous tokens.

To elaborate, write as $\bm Z$ the matrix of $N$ input tokens $\{\bm Z_{:,n}\}_{n=1}^N$ and as (\ref{eq:qkv}) the corresponding value, key, and query vectors  $\{\bm V_{:,n},\bm K_{:,n},\bm Q_{:,n}\}_{n=1}^N$. As discussed in the previous section, following associative memorization principles, the output token $\bm Y_{:,n}$ should generally attend to the value vector $\bm V_{:,n'}$ to an extent that depends on the corresponding query-key inner product $\bm Q_{:,n}^\top \bm K_{:,n'}$.  

To ensure that the most important values are further magnified in the weighted combination of values, the {softmax self-attention} mechanism leverages the \textbf{softmax function}. Given a vector of $N$ numbers $[z_1,...,z_N]$, here representing query-key inner products, the softmax function outputs $N$ non-negative numbers summing to 1, i.e., a probability distribution, given by \begin{equation}\text{softmax}_n(z_1,...,z_N)=\frac{\exp(z_n)}{\sum_{n'=1}^{N}\exp(z_{n'})},\end{equation} for all $n=1,...,N$. The softmax function can be thought of as a ``soft'' version of a maximum, in the sense that, as the gap between the largest score $z_n$ and the other scores $z_{n'}$ increases,  it tends to 1 for the index $n$, while approaching zero for all other indices $n'\neq n$.

Furthermore, to  allow flexibility in the way each token attends to other tokens,  softmax self-attention applies a \textbf{masking} mechanism (see Section \ref{sec:SSMfund}).  Specifically, given $N$ tokens to be processed, softmax self-attention defines a \textbf{mask} $M_{n,n'}$, which equals $0$ if token $n$ must attend to token $n'$ and while returning $-\infty$ otherwise. 

Given a mask $M_{n,n'}$, each $d$-th channel of the $n$-th output token is evaluated as the linear combination \begin{equation} 	Y_{d,n}= \sum_{n'=1}^N  S_{n,n'}  V_{d,n'}, \label{eq:selfatt}\end{equation} with attention scores \begin{equation}\label{eq:softmaxweights} S_{n,n'}=\text{softmax}_{n'}\left(\left\{ \frac{\bm Q_{n,:}^{\top}\bm K_{n'',:}+M_{n,n''}}{\sqrt{D_{\text{k}}}}\right\} _{n''=1}^{N}\right). \end{equation} Note that the inner products are scaled before being fed to the softmax function  in order to make their variance approximately independent of the dimension $D_\text{k}$.

 Unlike in the linear attention model (\ref{eq:ssm_sum_la1}), the  inter-token mixing operation (\ref{eq:selfatt}) implemented by softmax  attention  is not applied online over a computational time index $t$, but rather in a batch fashion, based on the entire input $\bm Z$.  This allows the attention score $S_{n,n'}$ to depend on {all} the keys $\{\bm K_{:,n'}\}_{n'=1}^N$ and not just on the key $\bm K_{:,n'}$. Furthermore, the mixed token value    (\ref{eq:selfatt}) may be a function of  all the values $\{V_{d,n'}\}_{n'=1}^N$ and not just on the ``previous'' values $\{V_{d,n'}\}_{n'=1}^n$. In particular,  when the mask $M_{n,n'}$ equals $0$ for $n'\leq n$  and $-\infty$ otherwise, one obtains a non-linear form of  autoregressive inter-token mixing as an alternative to linear attention (\ref{eq:ssm_sum_la1}). More generally, the attention scores (\ref{eq:softmaxweights}) can be strictly positive also for tokens $n'$ with $n'>n$, hence not being restricted to autoregressive mixing.

%
%

\subsection{Attention-Based NPEs}
As discussed, softmax self-attention  require the evaluation of the query-key inner products $\bm{Q}_{n,:}^\top\bm{K}_{n',:}$ in the attention scores  (\ref{eq:softmaxweights}), as well as the attention-value inner products in (\ref{eq:selfatt}). These inner products generally require the implementation of a number of multiply-and-accumulate operations that scales with the size of the vectors. Alternatively, as proposed in recent research, approximations of these inner products can be evaluated more efficiently by using encoding of the queries, keys, and values as sparse spiking signals along a virtual time axis. This approach forms the core of a number of  spiking transformers \cite{zhou2023spikformer, zhou2023spikingformer,abreu2025neuromorphic,song2025xpikeformer}. As illustrated in Figure \ref{fig:attentionsp}, the following main strategies have been proposed, which operate on binary spiking signals.

\begin{itemize}
    \item \textbf{LIF-based encoding with coincidence-based scores:} The linear projections (\ref{eq:qkv}) are first converted into spiking signals by passing each vector  through a LIF-based NPE. As a result, one obtains a spiking signal, defined along the virtual time axis $t$, per vector. The similarity between the two vectors is then  evaluated via a direct comparison of the corresponding binary vectors. To this end, a natural idea is to count the number of coincident spikes in the two vectors. This requires only a logical AND operation per entry to detect two spikes in the same position \cite{zhou2023spikformer, zhou2023spikingformer,abreu2025neuromorphic}. This metric is extremely cheap to evaluate, and it intuitively captures a notion of similarity encoded in the timings of the spikes in the two vectors. An alternative metric counts the  number of coincident binary digits -- thus including not only ones but also zeros. This metric requires an XNOR operation per entry of the encoded time series  \cite{xiao2025rethinking}. Figure \ref{fig:attentionsp} illustrates an example for spiking encodings corresponding to a key vector and a query vector. 
    \item \textbf{Stochastic computing:} The work~\cite{song2025xpikeformer} leverages   \textbf{stochastic computing} \cite{alaghi2013survey} to evaluate the inner products required by self-attention. Unlike the LIF-based encoding schemes used by the previous class of methods, stochastic computing recovers exact results as the virtual time index $t$ grows large. To this end,  queries, keys, and values are transformed into binary signals via Bernoulli encoding (recall (\ref{eq:Bernoulli})).  This representation enables multiplications to be performed using simple  AND gates. In fact, the result of the AND of two independent  Bernoulli random variables $\text{Bern}(p_1)$ and $\text{Bern}(p_2)$   is a Bernoulli random variable $\text{Bern}(p_1 p_2)$. Therefore, by the law of large numbers, as the virtual time index increases, the estimated rate of spike coincidences detected by the AND gates tends to the correct product value $p_1p_2$.
    \end{itemize}


	\section{Training Methods for Modern Neuromorphic AI}\label{sec:train}

	Training NPUs poses unique challenges due to their defining characteristics described in Section \ref{sec:charact}: \begin{itemize} \item  \textbf{Non-linear recurrence:} NPEs generally implement non-linear recurrent dynamics, where the non-linearity is introduced by the reset mechanism. This is problematic because it prevents the use of parallel computation strategies for supervised learning methods (see Section \ref{sec:parallel}). \item \textbf{Discrete activations:} The non-differentiable nature of the thresholding mechanisms underlying the generation of the outputs of NPEs renders the use of gradient descent methods challenging (see Section \ref{sec:surrogate}). \end{itemize}. This section briefly reviews training methodologies that address these challenges, from techniques based on teacher-forcing approaches and backpropagation (Section \ref{sec:parallel} and Section \ref{sec:surrogate}) to those leveraging reinforcement learning-like strategies and local adaptation (Section \ref{sec:reinforcement}). 

    \subsection{Backpropagation-Based Learning via Teacher-Forcing }\label{sec:parallel}

In a conventional supervised learning setting, the model is given examples of the form of a sequence of inputs $\bm x^T=\{\bm x[t]\}_{t=1}^T$. As shown in Figure \ref{fig:tokenization}, these may be obtained by tokenizing and embedding text or other types of data. Since the desired outputs $\bm x^T$ are given, this type of setting is also known as \textbf{teacher forcing}, as the ``teacher'' (the data) ``forces'' the desired behavior on the model \cite{williams1989learning}.  

In this case, the goal of learning is typically to minimize the discrepancy between the desired sequence $\bm x^T$ and the actual output of the model. Since the NPU's output is often of the form of a probability distributions over the vocabulary space,  the loss is typically chosen as the corresponding negative log-likelihood of the target $\bm x^T$. 

Denote the resulting training loss function as $\mathcal{L}(\theta,\bm x^T)$, where $\theta$ collects all the trainable parameters of the model (see (\ref{eq:general_dynamics})). Evaluating the gradient of this loss,  $\nabla_\theta \mathcal{L}(\theta,\bm x^T)$, requires backpropagating through the model's computational graph (see, e.g., \cite{simeone2022machine}). In general, NPEs -- and thus also NPUs -- are recurrent models implementing dynamics of the form (\ref{eq:lif_dynamics}). Therefore, evaluating the gradient $\nabla_\theta \mathcal{L}(\theta,\bm x^T)$ calls for the use of \textbf{backpropagation through time} (BPTT). The forward pass of  BPTT  goes sequentially through all time steps $t=1,2,...,T$; and  the backward pass follows the opposite order. This sequential processing may prevent the efficient training of models that operate on many tokens $T$ (apart from suffering from issues such as vanishing gradients).

    One of the key advantages of the SSM dynamics (\ref{eq:ssm_discrete}) is its  \textbf{linearity} in the state variables $\bm h[t]$. This makes it possible to avoid sequential operations during training via teacher forcing, leveraging instead parallel computations. To see this, consider first the case of constant gains $a[t]=a$ and $b[t]=b$. In such settings, the dynamics (\ref{eq:ssm_discrete}) can be expressed as a convolution between the impulse response of the system and the sequence of inputs $\bm k[t]v_d[t]$. Therefore, concatenating such inputs $\{\bm k[t]v_d[t]\}_{t=1}^T$ across $T$ time steps into a single vector $\bm z$ and, similarly, the corresponding output states $\{\bm h_d[t]\}_{t=1}^T$ into a vector $\bm h_d$, this convolution can be expressed as a matrix-vector product \begin{equation}\label{eq:Toeplitz} \bm h_d = \bm {C} \bm z, \end{equation} where $\bm{C}$ is a block-Toeplitz matrix dependent on the gains $a$ and $b$. 
    
    This matrix-product multiplication (\ref{eq:Toeplitz}) can be conveniently computed in \textbf{parallel} for all the entries of vector $\bm h_d$, or, more practically, across blocks of entries of this vector. In fact, given the input $\bm z$, each entry of the output $\bm h_d$ can be computed separately from the other entries by using the corresponding row of matrix $\bm C$. This parallelization implies that, with order $\mathcal{O}(T)$ parallel processors, one can evaluate the product (\ref{eq:Toeplitz}) in time of order $\mathcal{O}(T)$. This is a significant time complexity saving as compared to the sequential evaluation of the product (\ref{eq:Toeplitz}), which would take time of order $\mathcal{O}(T^2)$ to compute all output elements one by one.
    
     The total number of computations across all parallel processors can be further reduced from order $\mathcal{O}(T^2)$ to $\mathcal{O}(T)$ via the \textbf{parallel scan} algorithm, which is tailored to the partial-prefix sum form of the dynamics (\ref{eq:ssm_discrete}) \cite{blelloch1990prefix,gu2023mamba}. This algorithm has the added advantage that it can be also applied when the gains  $a[t]$ and $b[t]$ are not stationary due to implementation of gating (see Section \ref{sec:SSMfund}). However, in some cases, it may be less preferable for hardware deployments that can benefit from well-optimized kernels for matrix multiplication.
    
As discussed in Section \ref{sec:real}, due to the presence of the reset mechanism, the LIF-based dynamics (\ref{eq:lif_dynamics}) cannot be expressed as an SSM, calling for a \textbf{sequential} evaluation of the gradient via BPTT. However, having removed the reset mechanism, the LIF dynamics can be expressed as a convolution, since the LIF model (\ref{eq:lif_dynamics}) with matrix $\bm W_\text{res}=\bm 0$ is linear in the state variables $\bm u[t]$ as for SSMs. Therefore, the evaluation of the membrane potential dynamics can be implemented in parallel through convolutions or the parallel scan algorithm as discussed above  \cite{fang2023parallel,bal2024pspikessm}.
	
	Retaining the reset mechanism while enabling parallelization requires approximation techniques. For example, one can train a neural network that predicts the spike sequence and then use the predictions in lieu of the true spikes for parallel computation~\cite{du2024spikingssms}. An alternative iterative mechanism to obtain the spiking times was discussed in \cite{zhong2024spike}. An event-driven parallel scan approach is studied in \cite{tang2026spiky}.

	\subsection{Gradient Evaluation via Surrogate Gradients}\label{sec:surrogate}
	
    While removing the reset mechanism reduces the complexity of the forward and backward passes, the gradient of the training loss for an NPU will result in an all-zero tensor for almost all values of the parameters $\theta$. In fact,  small changes in parameters $\theta$ generally do not modify the discrete activations of the NPEs, as they do not affect the output of the threshold functions in (\ref{eq:spike_generation})-(\ref{eq:gif_spike}). 

    To obviate this problem, one approach is the \textbf{straight-through estimator}, which replaces the threshold functions with  the identity function during the backward pass \cite{bengio2013estimating}. A potentially more effective approach is to use as a replacement  a smooth function that serves as a closer approximation, or \textbf{surrogate}, of the original threshold function \cite{neftci2019surrogate}. For example, for the case of binary outputs (\ref{eq:spike_generation}), common choices of approximating functions include the sigmoid function, the arctangent function, and piece-wise linear approximations \cite{bellec2018long,zenke2018superspike}.

	
	

	\subsection{Local Learning via Feedback-Based Local Updates}\label{sec:reinforcement}

    Backpropagation may be prohibitively complex and incompatible with some neuromorphic computing platforms \cite{davies2021advancing}. Furthermore, it is not applicable to control settings in which data cannot ``force'' desired behavior. In fact, in such scenarios, the model takes actions and is only then rewarded or penalized   \cite{fremaux2013reinforcement}. For all these reasons, it is important to consider alternative learning paradigms based on feedback and local adaptation, rather than on backpropagation and teacher forcing objectives. Most of the work on this topic has focused so far on standard SNN models defined by networks of spiking neurons (see Section \ref{sec:charact}), and we limit the discussion here to such settings.

   The most fundamental local learning rule is \textbf{spike-timing-Dependent plasticity} (STDP), which adjusts weights connecting spiking neurons based on the relative timing of pre- and post-synaptic spikes. STDP is inspired by neuroscientific principles, and operates in a fully unsupervised way \cite{dan2004spike}.  

   However, STDP alone is insufficient for supervised, or goal-directed,  learning. \textbf{Three-factor learning rules} extend STDP by incorporating a global reward or error signal, enabling learning from outcomes~\cite{gerstner2018eligibility}. Such rules can be derived from first principles for spiking models with probabilistic binary spiking mechanisms as in (\ref{eq:prob})~\cite{jang2019introduction}. In this type of models, neurons respond probabilistically to their inputs. The loss function evaluated at the last layer serves as a reward signal to encourage or discourage the realized behavior on the basis of adherence to labeled data or control goals. 
	
	
	
	
	\section{Conclusion}\label{sec:conclusions}

	This paper has provided an overview of modern neuromorphic AI through the lens of intra-token and inter-token processing. Early neuromorphic systems, primarily based on SNNs, focused on intra-token processing tasks such as image classification by leveraging sparse communication along a virtual time axis. More recently, neuromorphic principles have been extended to inter-token processing, enabling efficient sequence modeling through both state-space dynamics and sparse self-attention mechanisms. These developments have been mirrored by advances in traditional mainstream AI architectures, particularly transformers and SSMs, which have incorporated neuromorphic principles such as integer activations. 
	
	Despite significant progress, several important research challenges remain open. First, the definition and evaluation of neuromorphic language and multi-modal models incorporating the latest advances in state-space models, such as selective SSMs with adaptive gating mechanisms and matrix-based gains, requires further investigation. Second, the systematic reduction of activation precision in state-of-the-art LLMs remains an open challenge. While recent work has demonstrated LLMs with 4-bit activations, the path toward fully spike-based language models with binary or ternary activations, while maintaining competitive performance, may require further algorithmic innovations in both model architecture and training procedures. 
	
	Third, with an eye to hardware implementations, the design and evaluation of zeroth-order learning methods for memory-efficient training of spiking transformers may represent a useful direction. Such methods  may enable on-device learning and adaptation without the memory overhead of BPTT, making neuromorphic AI more practical for edge deployment. More broadly, developing \textbf{hardware-software co-design} methodologies that fully exploit the capabilities of emerging neuromorphic processors for both intra-token and inter-token processing, including analog in-memory computing, is a critical research direction.

	As neuromorphic AI systems increasingly converge with mainstream architectures, interdisciplinary collaboration between neuroscience, machine learning, and hardware engineering will be essential. The principles that have guided neuromorphic computing -- sparsity, event-driven computation, local learning, and stateful dynamics -- could ultimately turn out to be indispensable to optimize the intelligence-per-joule ratio.

%
%
	
	\bibliographystyle{IEEEtran}
\bibliography{refs}
	
\end{document}